# Biologically Inspired Spiking Neurons:
# Piecewise Linear Models and Digital Implementation

Hamid Soleimani, Arash Ahmadi *Member* IEEE and Mohammad Bavandpour

*Abstract*— there has been a strong push recently to examine biological scale simulations of neuromorphic algorithms to achieve stronger inference capabilities. This paper presents a set of piecewise linear spiking neuron models, which can reproduce different behaviors, similar to the biological neuron, both for a single neuron as well as a network of neurons. The proposed models are investigated, in terms of digital implementation feasibility and costs, targeting large scale hardware implementation. Hardware synthesis and physical implementations on FPGA show that the proposed models can produce precise neural behaviors with higher performance and considerably lower implementation costs compared with the original model. Accordingly, a compact structure of the models which can be trained with supervised and unsupervised learning algorithms has been developed. Using this structure and based on a spike rate coding, a character recognition case study has been implemented and tested.

*Index Terms*— Spiking Neural networks, Piecewise Linear Model, Field Programmable Gate Array (FPGA), Spike Rate Learning.

## I. INTRODUCTION

SPIKING Neural Networks (SNN) have received a considerable attention in the artificial neural network community during the past few years, due to their behavioral resemblance to biological neurons. Motivated by biological discoveries, pulse-coupled neural networks with spike-timing are considered as an essential component in biological information processing systems, such as the brain. Accordingly, many different models have been presented for spiking neural networks to reproduce their dynamical behavior. These models are based on the bio-chemical inspection of the neuron structures and mostly are expressed in the form of differential equations. Although detailed neuron models can imitate most experimental measurements to a high degree of accuracy; due to their complexity, most of them are difficult to be used in large scale artificial spiking neural networks [1],[2]. Therefore, varieties of simplified models are presented for studies in the field of neural information coding, memory and network dynamics.

In general, there is a tradeoff between model accuracy and its computational complexity. For instance, when it is required to understand how neuronal behavior depends on measurable physiological parameters, such as the maximal conductance, steady state activation/inactivation functions and time constants, the Hodgkin–Huxley type [1] models are more suitable, which are computationally expensive and cannot be simulated in large numbers. On the other hand, if the goal is to understand temporal behavior of the cortical spike trains or spike-timing to investigate how the mammalian neocortex processes information; spike-based models are appropriate, which can exhibit biological neuron signaling properties [3].

Izhikevich in [4] has introduced one of the widely accepted models, which can reproduce a variety of neuron firing patterns. This model is claimed to be one of the simplest possible models that can exhibit all the firing patterns. This neuron model has been commonly accepted as an accurate and computationally affordable model yet producing a wide range of cortical pulse coding behaviors.

Implementation of these models, targeting different platforms, has been subject of studies in terms of efficiency and large scale simulations based on optimal transfer capability of the spike signals provided by address event representation [5],[6]. There exist three major approaches for this challenge:

1) Analog implementations are considered to become a strong choice for direct implementation of neuro-inspired systems [7]-[12]. In this approach, electronic components and circuits are utilized to mimic neurological dynamics. Due to its high performance and well developed technology, an analog VLSI implementation enables prototyping of neural algorithms to test theories of neural computation, structure, network architecture, learning and plasticity and also simulation of biologically inspired systems in a real-time operation. This is of particular interest for sensory processing systems and biologically-inspired robotics. Although these analog solutions are fast and efficient, they are inflexible and require a long development time [13],[14].

2) Special purpose hardware have been developed to implement neurobiological functions using software based systems for large scale simulations. Examples are Blue Brain [15], Neurogrid [16] and SpiNNaker [17]. Even though these systems are flexible and biologically realistic with considerably high performance, the presented hardware approaches suffer from limited programmability and high-cost. Unfortunately the cost and development time make these approaches impractical for public access, general purpose large-scale simulations.

3) Recently, reconfigurable digital platforms have been used to realize spiking neurons [18]-[32]. This approach uses digital computation to emulate individual neural behaviors in parallel and distributed network architecture to implement a system level dynamic. Although digital computation consumes more silicon area and power per function in comparison with the analog counterpart, its development time is considerably lower and is not susceptible to power supply, thermal noise or device mismatch. In addition, high precision digital computation makes it possible to implement networks with high dynamic range, greater stability, reliability and repeatability.

Recently studies have been published [22]-[27], which have implemented the Izhikevich neuron model instead of the Integrate and Fire (IF) model on FPGAs. La Rosa et al. [28] and Fortuna et al [29] simulated neuron networks on an FPGA in which the primary objective was to examine the feasibility of FPGA implementations of the model and to show that hardware can reproduce a wide range of neural responses. Mokhtar et al. [30] simulated 48 neurons based on the Izhikevich model on a Virtex II Pro FPGA for maze navigation and Cassidy et al. [31],[32], implemented an FPGA based array consisting of 32 physical neurons. It should be noted that the main limitation of the previously published works to implement large scale networks on FPGA is the number of available fast multipliers on the chip. For instance, in [31] and [32] only 32 neurons are implemented because there are only 32 embedded multipliers in the utilized FPGA boards. It is notable that a multiplier is an expensive building block in terms of area, latency and power consumption.

This paper presents a set of PWL multiplier-less models, which are modifications of the Izhikevich model. The proposed models are efficiently implementable in both analog and digital platforms for large scale simulation projects. These models use the same approach as the original model with a modification by which the "square" operation in the specification equation is replaced with a "comparison" or "absolute value" operations both of which are far less expensive compared to the "square" function in either analog or digital implementation. From a digital implementation point of view, this modification simplifies required hardware for the model by replacing "multiplication" with "addition" and "logic shift", which makes it possible to realize a large number of neurons on a single FPGA board. Digital implementations on FPGA show hardware cost of the proposed model is considerably lower yet demonstrate similar dynamic behavior as the original one.

To investigate network behavior of the PWL neuron models, a pattern recognition network is trained using a rate-based algorithm. Results show suitability of this algorithm for training neurons with the Izhikevich model. Although in [33] back propagation rule [34] has been used to train a SNN, their approach contains multi sub-synapse, which requires rather large area in implementation compared with the proposed training algorithm. In addition, their data coding scheme is difficult for hardware implementation compared with the presented training method.

The paper is organized as follows: The next section presents a brief background of the original model, while section III discusses the proposed neuron models. Finding coefficients of the new models based on an errors assessment approach is presented in the section IV. Simulation based dynamic analyses of the models are offered in Sections V. Network behavior of the models and hardware design details are explained in section VI and VII respectively. A pattern recognition case study is presented in section VIII and implementation results are in section IX.

## II. Background

By generating sequences of action potentials, neurons process and encode information. Neurons encode computations into sequences of spikes which are biophysically determined by the cells' action potential generating mechanism. Izhikevich in [4] and [35], proposed a model which consists of two coupled differential equations as:

$$\begin{cases} v' = 0.04v^2 + 5v + 140 - u + I \\ u' = a(bv - u) \end{cases} \quad (1)$$

with the auxiliary reset equations:

$$v \geq v_{th} \qquad then \qquad \begin{cases} v \leftarrow c \\ u \leftarrow u + d \end{cases} \quad (2)$$

where **v** represents the membrane potential of the neuron and **u** represents a membrane recovery variable, which accounts for the activation of K+ ionic currents and inactivation of Na+ ionic currents and it provides negative feedback to **v**. After the spike reaches its apex ($v_{th}$), the membrane voltage and the recovery variable are reset according to the equations above. If **v** goes over $v_{th}$, it first resets to $v_{th}$, and then to **c** so that all spikes have equal magnitudes. The part $0.04v^2 + 5v + 140$ is chosen so that **v** is in mv scale and time is in ms. Although this is known as the most practical accurate model; still there are several challenges in realizing the model on digital or analog circuits. The difficulty of implementation arises from the quadratic part of the model which is shown by the parabolic curve in Fig. 1-a.

## III. Modified Neuron Models

In this section, to improve computational efficiency of the Izhikevich model, three piecewise linear approximations are proposed.

### A. Second order piecewise linear model

As it is shown in Fig. 1-b, the proposed second order piecewise (2PWL) model approximates the quadratic part of the Izhikevich model with two crossed lines. This approximation can be formulated as:

$$\begin{cases} v' = k_1 |v + 62.5| - k_2 - u + I \\ u' = a(bv - u) \end{cases} \quad (3)$$

This approximation provides two degrees of freedom to achieve the closest behavior to the original model.

### B. Third order piecewise linear model

For third order piecewise (3PWL) approximation the following function is presented:

$$\begin{cases} v' = k_1 (|v + 62.5 + k_2| + |v + 62.5 - k_2|) - k_3 k_2 k_1 - u + I \\ u' = a(bv - u) \end{cases} \quad (4)$$

This new nonlinear function is depicted in Fig. 1-c. As can be seen, this approximation provides three degrees of freedom to achieve the closest behavior to the original model.

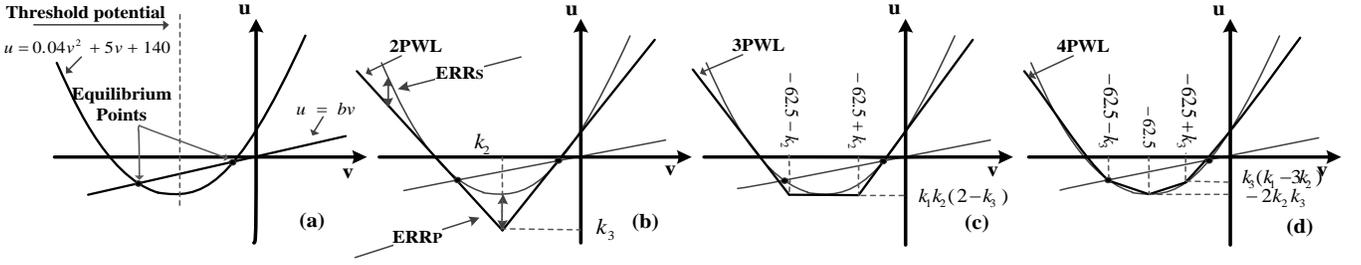

Fig. 1. Equilibrium **u-v** locus of the neuron models and their corresponding k coefficients a) Izhikevich neuron model. b) Second order piecewise linear model, c) Third order piecewise model. d) Forth order piecewise model.

Compared to the 2PWL model, this model has advantages and disadvantages. In terms of implementation, the 3PWL approximation is more expensive compared to the 2PWL, but the behavior of 3PWL model can be closer to the original model by appropriate choice of the coefficients

### C. Forth order piecewise linear model

The proposed forth order piecewise (4PWL) approximation is formulated as:

$$\begin{cases} v' = k_2 \left( \left| v + 62.5 + k_3 \right| + \left| v + 62.5 - k_3 \right| \right) + k_1 \left| v + 62.5 \right| - 4k_2 k_3 - u + I \\ u' = a(bv - u) \end{cases} \quad (5)$$

where $k_1$, $k_2$ and $k_3$, similar to the other PWL models, are constant values. This new nonlinear function is depicted in Fig. 1-d. As seen, this approximation provides three degrees of freedom for achieving the closest behavior to the original model. This model requires more complex circuit implementation compared to the other PLW models, but has a very close behavior compared to the other proposed models.

In these models, k coefficients ($k_1$, $k_2$ and $k_3$) are constant values, which must be pre-calculated. The values of the k coefficients are chosen based on two basic aspects: model accuracy and implementation simplicity. In terms of accuracy, the error minimization method is explained in section IV where for the model accuracy, we have to bear in mind that $v \geq v_{th}$, which means that $v$ and $u$ are practically bounded and this approximation just needs to be valid within these limits. To simplify the implementation, we should choose values for k coefficients, which can be implemented using only "logic shift" and "add".

### IV. FINDING k COEFFICIENTS FOR THE PWL MODELS

Since the dynamic behavior of a neural network depends on both the network structure as well as the model of the single neuron, the accuracy and correctness of the proposed PWL models need to be examined in both single neuron along with populations of neurons in a network.

To evaluate single neuron dynamic, one needs to understand the relationship between neuron behavior and its equilibrium locus, as depicted in Fig. 2. Fig. 2-a and b show different phase state paths between threshold line and steady state curve of the **v**. As it is observable, the refractory phase in membrane potential response strongly depends on the slope of the quadratic part of the graph (refractory phase is the amount of time it takes for an excitable membrane to be ready for a second stimulus once it returns to its resting state following excitation). Since in PWL models this quadratic part is replaced with linear approximations, refractory response of the PWL models must be examined for any affection.

Another point, which needs to be taken into account, is the return path of the membrane potential to the **u** line after neuron excitation, as shown in Fig. 2-a. During this phase, the membrane potential rises up to the peak point. The curvature and smoothness of this path affects the excitation form of the membrane potential response [36]. Moreover, the length of this path affects the firing rate frequency in the dynamic behavior. It means that the smaller path, the increasing firing rate will be. Based on these initial reviews error values are defined as follows.

**ERR_S:** This error can be defined as the difference between the slope of the main curve and the PWL models as shown in Fig. 1-b. This error affects the refractory phase response, where the bigger deference makes the refractory phase response slower. This error can be formulated as:

$$ERR_S = \left| \left( \frac{du}{dt} \right)_{\text{Original model}} - \left( \frac{du}{dt} \right)_{\text{PWL models}} \right| \quad (6)$$

**ERR_S** for different PWL models are presented in Table I. Since this error is more important in the points where neuron moves along the excitation path, the slope error in Table I are presented as functions of $v$.

**ERR_P:** This error is defined as the difference between the main curve and PWL models at the lowest point of the curves as depicted in Fig. 1. The PWL models may shift peaks at the bottom of the curves. This determines the initial excitation for the neuron, it means, when this error is bigger the neuron requires bigger input stimulus (I) to put the curve in a suitable place for excitation in the **u-v** plane. In addition, this error has a strong effect on the excitation form of the membrane potential (**v**). Results are presented in the Table I.

TABLE I. Error formulations for PWL models.

|  | 2PWL | 3PWL | 4PWL |
|---|---|---|---|
| **ERR_S** | $\|0.08v + 5 - k_1\|$ | $\|0.08v + 5 - 2k_1\|$ | $\|0.08v + 5 - (2k_2 + k_1)\|$ |
| **ERR_P** | $\|k_2 - 16.25\|$ | $\|k_1 k_2 (2 - k_3) - 16.25\|$ | $\|2k_3 k_3 - 16.25\|$ |

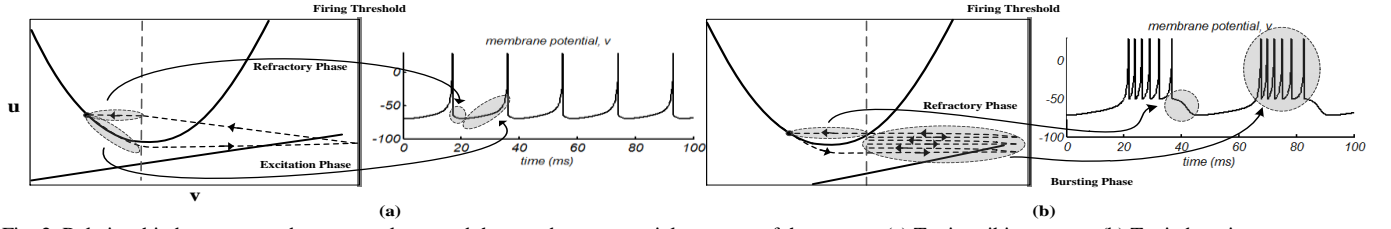

Fig. 2. Relationship between steady state **v-u** locus and the membrane potential response of the neuron. (a) Tonic spiking neuron (b) Tonic bursting neuron.

According to the discussions above, an error minimization algorithm is proposed for optimizing k coefficients of the PWL models. This method finds a series of coefficients based on comparisons between the original model and the PWL models for each type of neurons. There are different options for the Cost Function (CF) in this optimization method. In [37], spikes or rates of the neuron behavior have been considered using Gamma CF for optimization, where in [38] an analog implementation based on memristor crossbar structure is presented in which spike shaping for STDP learning is considered. In the proposed search method we use a CF based on both rate and spike shape for N points of the **v** signal as:

$$CF = \frac{1}{N} \sum_{i=1}^{N} \frac{(v_{origin}(i) - v_{PWL}(i))^2}{v_{origin}^2(i)} \qquad (7)$$

The proposed algorithm is presented in Fig. 3. In line 1 the initial value for k coefficients and the other variables like **CF_Temp** are assigned. In this algorithm, there are 3 main loops. The first loop is in the line 2. This loop is repeated until P point which P is:

$$P = \frac{k_3}{\Delta k_3} \qquad (8)$$

where $\Delta k_3$, is the $k_3$ increment step in each loop. In each loop for every new $k_3$, algorithm resets **v** (membrane potential of the neuron) and **u** (recovery variable). Similarly, Q and R are:

$$Q = \frac{k_2}{\Delta k_2} \quad \text{and} \quad R = \frac{k_1}{\Delta k_1} \qquad (9)$$

where the value of $\Delta k_2$ and $\Delta k_1$ are the increment steps of $k_2$ and $k_1$ respectively. All the feasible values for k coefficients are checked within these three loops. The other loop in line 5 is used to delete the unstable and noisy part of the neurons signal which must be ignored for sampling and comparing models output. The valid part of the output signal is called **S**. There is another point to be considered for a fair comparison of the neurons signaling, which is the synchronization of the outputs. There are two **While loops** which pause signals in their spike instant. After synchronization, the CF is computed for N points of the signals. For reliability of the error analysis the value of N should cover M cycles of the signals. Therefore, **CF_Temp** is accumulated for M repetitions and the average is calculated in line 11.

// assign initial values for k coefficients and the other variables.
1: $k_1$=M, $k_2$=N, $k_3$=O, CF_Temp=0;
2: **For** P point **do**{
Reset v & u;
3: **For** Q points **do** {
Reset v & u;
4: **For** R points **do**{
Reset v & u;
CF_Temp =0;
5: //Call the original model and PWL models and ignore the unstable part of the neuron behavior before S part.
**For** points in S **do**{
6: Original model Function()
PWL models()
Checking auxiliary reset equation.}
**For** T points **do**{
7: // synchronization of the original model and PWLs.
**While** (v>v_th) **{**
8: Original model Function()}
**While** (v>v_th) **{**
9: PWL model Function()}
**For** N points (M cycles ) **do** {
10: // call original model and PWL model.
CF_Temp =CF_Temp + $(V_{origin} - V_{PWL})^2/(V_{origin})^2$ ;}
CF=(CF_Temp /N);
11: $k_1$=$k_1$+$\Delta k_1$.}
$k_1$=M; $k_2$=$k_2$+$\Delta k_2$;}
$k_1$=M; $k_2$=N; $k_3$=$k_3$+$\Delta k_3$}
12: **end**

Fig 3. The pseudo code of the search algorithm for error assessment.

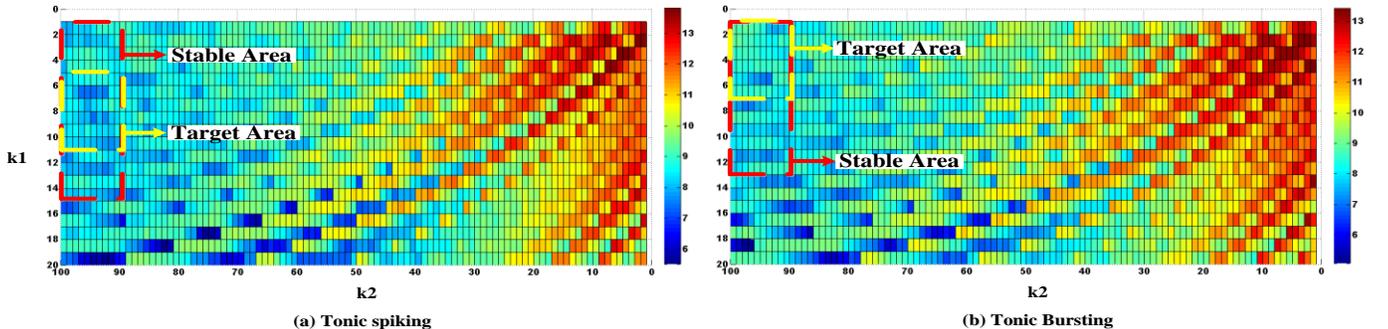

Fig 4. Color graphs for k coffcients with normalized axes. There is a low error zone (cold color) in the plot in where CF is consistently low  and also there is a target area in this place where the k coefficients are digital (fixed-point numbers). The CF is ploted for (a) tonic spiking and (b) tonic bursting neurons.

TABLE II. The optimized k coefficients for PWL models based on error assessment procedure.

| Neuron Type | 2 PWL | | | | 3 PWL | | | | | 4 PWL | | | | |
|---|---|---|---|---|---|---|---|---|---|---|---|---|---|---|
| | $ERR_p$ | ERR% | $K_1$ | $K_2$ | $ERR_p$ | ERR % | $K_1$ | $K_2$ | $K_3$ | $ERR_p$ | ERR% | $K_1$ | $K_2$ | $K_3$ |
| Tonic spiking | 3.75 | 8.5 | 0.75 | 20 | 0.3 | 6.6 | 0.625 | 5.8 | 6.4 | 0.25 | 1.8 | 0.375 | 0.75 | 11 |
| Phasic spiking | 1.75 | 7.4 | 0.5 | 18 | 0.3 | 6.4 | 0.625 | 5.8 | 6.4 | 0.25 | 1.5 | 0.375 | 0.75 | 11 |
| Tonic bursting | 3.75 | 10.8 | 0.625 | 20 | 0.3 | 6.2 | 0.625 | 5.8 | 6.4 | 0.25 | 1.9 | 0.375 | 0.75 | 11 |
| Phasic bursting | 3.75 | 9.2 | 0.5 | 20 | 0.5 | 5.1 | 0.5 | 7 | 6.5 | 0.25 | 1.4 | 0.375 | 0.75 | 11 |
| Mixed mode | 1.75 | 8.2 | 0.5 | 18 | 0.5 | 4.9 | 0.5 | 7 | 6.5 | 0.25 | 1.5 | 0.375 | 0.75 | 11 |
| Spike frequency adaptation | 1.75 | 7.8 | 0.375 | 18 | 0.5 | 5.1 | 0.5 | 7 | 6.5 | 0.25 | 1.4 | 0.375 | 0.75 | 11 |
| Class 1 | 1.75 | 8.3 | 0.375 | 18 | 0.5 | 5.2 | 0.5 | 7 | 6.5 | 0.25 | 0.7 | 0.375 | 0.75 | 11 |
| Class 2 | 1.75 | 10.1 | 0.625 | 18 | 0.5 | 4.4 | 0.5 | 7 | 6.5 | 0.25 | 0.7 | 0.375 | 0.75 | 11 |
| Spike latency | 1.75 | 8.3 | 0.625 | 18 | 0.5 | 6.1 | 0.5 | 7 | 6.5 | 0.25 | 1.3 | 0.375 | 0.75 | 11 |
| Subthreshold oscillations | 1.75 | 9.2 | 0.875 | 18 | 0.5 | 5.5 | 0.5 | 7 | 6.5 | 0.25 | 1.9 | 0.375 | 0.75 | 11 |
| Resonator | 1.75 | 9.8 | 0.875 | 18 | 0.5 | 4.9 | 0.5 | 7 | 6.5 | 0.25 | 1.3 | 0.375 | 0.75 | 11 |
| Integrator | 1.75 | 8.7 | 0.875 | 18 | 0.5 | 6.3 | 0.5 | 7 | 6.5 | 0.25 | 1.1 | 0.375 | 0.75 | 11 |
| Rebound spike | 1.75 | 7.5 | 0.875 | 18 | 0.5 | 3.4 | 0.5 | 7 | 6.5 | 0.25 | 1.6 | 0.375 | 0.75 | 11 |
| Rebound burst | 1.75 | 8.1 | 0.375 | 18 | 0.5 | 5.1 | 0.5 | 7 | 6.5 | 0.25 | 1.7 | 0.375 | 0.75 | 11 |
| Threshold variability | 1.75 | 9.9 | 0.375 | 18 | 0.5 | 5.8 | 0.5 | 7 | 6.5 | 0.25 | 0.9 | 0.375 | 0.75 | 11 |
| Bistability | 5.75 | 10.9 | 2 | 22 | 1.75 | 6.1 | 1.25 | 12 | 3 | 0.25 | 0.9 | 0.375 | 0.75 | 11 |
| Depolarizing after-potential | 1.75 | 7.4 | 0.625 | 18 | 0.5 | 6.1 | 0.5 | 7 | 6.5 | 0.25 | 0.8 | 0.375 | 0.75 | 11 |
| Accommodation | 1.75 | 10.8 | 0.625 | 18 | 0.5 | 4.4 | 0.5 | 7 | 6.5 | 0.25 | 1.1 | 0.375 | 0.75 | 11 |
| Inhibition-induced spiking | 1.75 | 7.9 | 0.625 | 18 | 0.5 | 3.2 | 0.5 | 7 | 6.5 | 0.25 | 0.7 | 0.375 | 0.75 | 11 |
| Inhibition-induced bursting | 1.75 | 8.5 | 0.625 | 18 | 0.5 | 5.1 | 0.5 | 7 | 6.5 | 0.25 | 0.5 | 0.375 | 0.75 | 11 |
| Mean Error % | 2.25 | 8.865 | - | - | 0.5325 | 5.295 | - | - | - | 0.25 | 1.235 | - | - | - |

Table II presents k coefficients for the most important neuron types of the Izhikevich model. As it can be seen, more complex PWL models (models with more number of lines 4>3>2) are more consistent with optimum k coefficients. In other words, 4PWL model can optimally create different neural behaviors with the same $k_1$, $k_2$ and $k_3$, which is not possible in 3PWL and 2PWL. This is because of the nature of the models and the freedom degrees they offer in k coefficients.

As it is expected, optimum values for k coefficients are not unique. Therefore, other issues such as hardware implementation can be used to narrow down the options. To have a better perception, error values are shown for different k coefficients in a color graph in Fig. 4. This graph is plotted for 2PWL model in tonic spiking and tonic bursting neuron types. The range of $k_1$ is 0.1-8 by steps size of 0.01 and $k_2$ is 10-25 by step size of 1.

From hardware implementation point of view, in addition to the minimum value of the error, error variations around the minimum point is also important. k coefficients and all related calculations in the model need to be realized using digital arithmetic. Limited precision of the digital systems (i.e. n-bit fixed-point numbers) as well as optimization techniques, which designer might apply, make it crucially important to choose k coefficients within regions in which error is low for all the points. In other words, for k coefficients instead of 'minimum point' we should look for a 'minimum zone' (Stable Area). As it can be seen in Fig. 4, there is a minimum zone (left upper- hand corner of each Figure) in which error is low in average and satisfies digital implementation limitations (Target Area).

## V. VARIOUS NEURON-LIKE RESPONSES

Based on simulation analysis, in this section we show that the PWL models can exhibit various neuron-like responses. Fig. 5 shows time waveforms of the original and the PWL models for various input current I. In this figure, for the original model and the PWL models, the input current is increased gradually until transition occurs. As a result, the response of the PWL models changes from the resting state to the tonic spiking via the tonic bursting as the input current (I) increases.

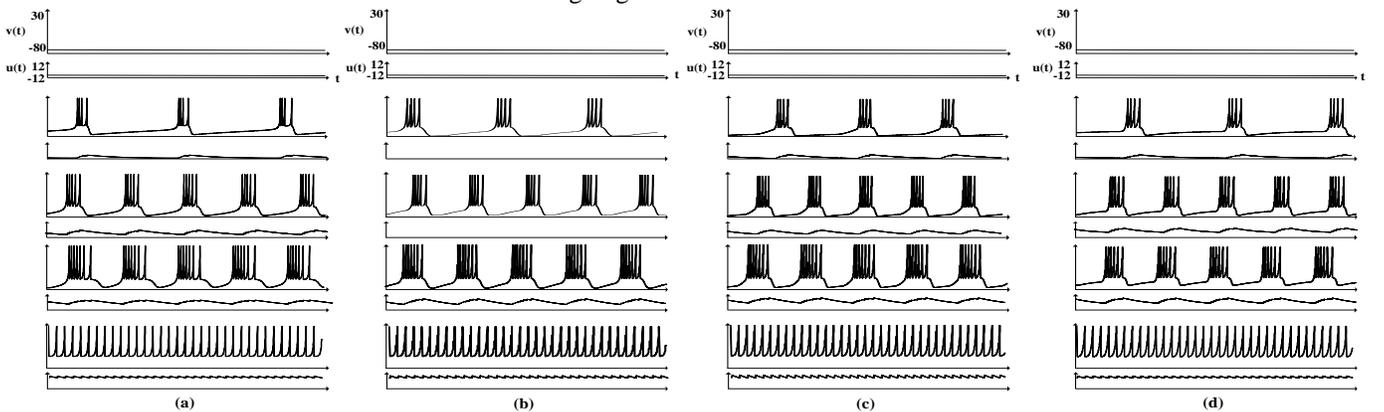

Fig 5. The "resting ↔ tonic bursting ↔ tonic spiking" transitions for the original and PWL models. (a) The original model with increasing input current as 0−>4.5−>12.5−>19.5 (b) The 4PWL model with increasing input current as 0−>4.5−>12.5−>19.5 with K=(0.375,0.75,11) (c) The 3PWL model with increasing input current as 0−>4.5−>12.5−>19.5 with K=(0.625, 5.8, 6.4) (d) The 2PWL model with increasing input current as 0−>5.5−>14−>22 with K=(0.625, 20).

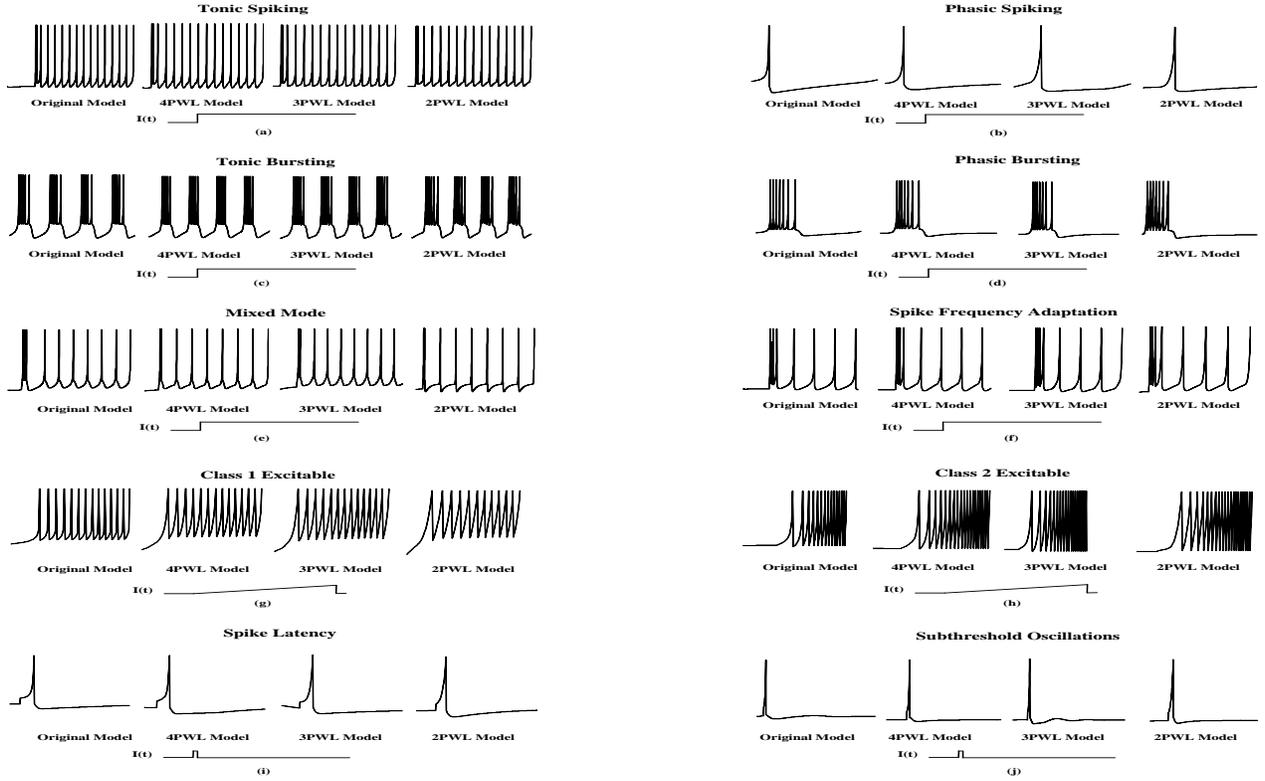

Fig 6. Comparisons between the PWL and Izhikevich (Original) models response to the input I(t). (a) Tonic spiking (b) Phasic spiking (c) Tonic bursting (d) Phasic bursting (e) Mixed mode (f) Spike frequency adaption (g) Class 1 excitable (h) Class 2 Excitable (i) Spike latency. (j) Sub-threshold oscillations.

These transitions are similar to the "resting ↔ tonic bursting ↔ tonic spiking" behavior as observed in biological neurons [35] and it shows that the PWL model can mimic these behaviors like the original model. It should be noted that these transitions are occurred based on bifurcation phenomena. In addition, the PWL model can exhibit various biologically neuron responses. For instance, Fig. 6-a to j show 10 types of responses of the PWL neuron models to an input I, compared with the Izhikevich model [3], [35]. As it is shown in this figure, 4PWL model has the closest behavior to the original model. It should be noted that the bifurcation analysis can help better understanding and mathematical proving of the proposed models, however since the main scope of the paper is not mathematical investigation and bifurcation analysis of the neural behavior, this subject has been scheduled for future works.

## VI. NETWORK BEHAVIOR

Precise spike timing is an important factor in SNN studies. For example, in spike time dependent plasticity the spike timing is the most important parameter in learning [39]. Therefore, to investigate the network dynamics of the neurons with the proposed PWL models and compare it with the original neuron model, a network of randomly connected 2000 neurons is simulated. Motivated by the anatomy of a mammalian cortex, we choose the ratio of excitatory to inhibitory neurons to be 4 to 1 and make the inhibitory synaptic connections stronger. Besides the synaptic inputs, each neuron receives a noisy thalamic [4]. The raster plots of the simulations are presented in Fig. 7. The network activities of the original model and the proposed PWL models with the approximately same inputs are very similar in structure, but differ in the precise details; however, they show the same rhythm of 5Hz. Since the statistical nature of the neural behavior is generally of interest, these differences may not be significant. For better understanding the differences between original and proposed models in network behavior, an error criterion is defined based on Relative Error (RE). This error was applied to the PWL models in each spike firing then mean value over 1000 ms has been calculated (MRE). This error can be formulated as:

$$\text{MRE}_{2PWL}\% = \frac{\left|\frac{\Delta t_{2PWL_1}}{t_{o_1}}\right| + \left|\frac{\Delta t_{2PWL_2}}{t_{o_2}}\right| + \cdots\cdots}{N} \times 100 = \frac{\sum_{i=1}^{N}\left|\frac{\Delta t_{2PWL_i}}{t_{o_i}}\right|}{N} \times 100 \quad (10)$$

where $\Delta t_{2PWL_i}$ is time difference between $i_{th}$ spike in PWL and original model as depicted in Fig.8. This procedure is applied for the all types of Izhikevich neuron models. Table III presents the MRE of randomly connected 2000 neurons for PWL models. As we expected, results show 4PWL model has the best performance where the 3PWL and the 2PWL models place second and third respectively. We should keep in mind that the major reason of these differences between PWL models and the original model (especially 2PWL model) in dynamic state is ERR$_P$. As it is mentioned in section IV, this error affects the excitation in the proposed models especially in the 2PWL, which results in a change in the frequency of neurons firing rate. So for getting the best performance from the PWL models, I(t) should be regulated accurately. The results in Fig. 7 certainly will be better with more accurate regulating initial input in all proposed PWL models.

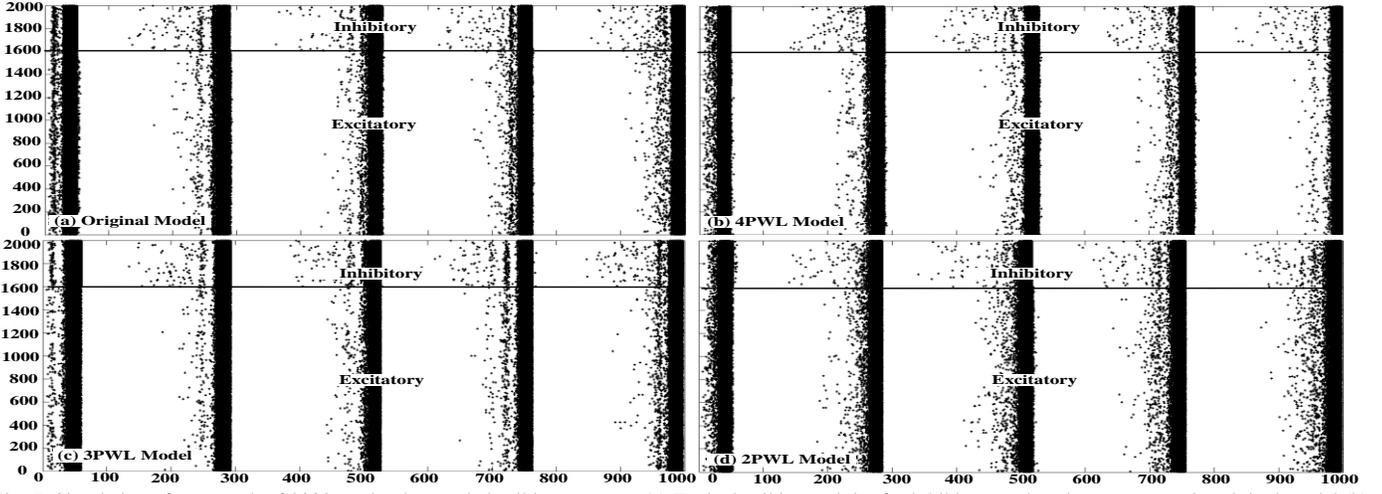

Fig. 7. Simulation of a network of 2000 randomly coupled spiking neurons. (a) Typical spiking activity for inhibitory and excitatory neuron in original model (b) Typical spiking activity for inhibitory and excitatory neuron in 4PWL model (c) Typical spiking activity for inhibitory and excitatory neuron in 3PWL model (d) Typical spiking activity for inhibitory and excitatory neuron in 2PWL model. All spikes were equalized at +30 mV by resetting v first to +30 mV and then to c.

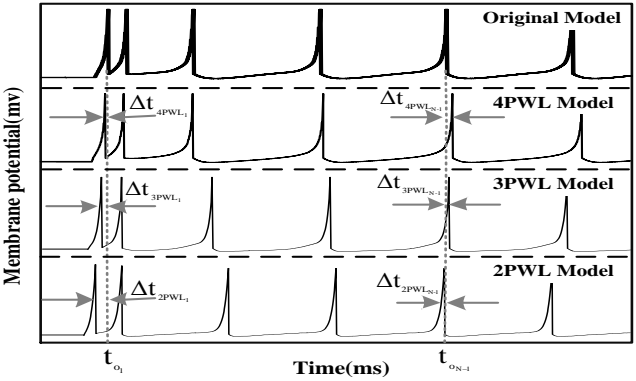

Fig. 8. Spike firing plot of PWL and original models. This plot shows the time differences of the spikes in PWL models compared to the original model.

TABLE III. The mean relative error for all PWL models in dynamic system is presented.

| Neuron type | 2PWL MRE% | 3PWL MRE % | 4PWL MRE% |
|---|---|---|---|
| Tonic spiking | 5.14 | 3.45 | 1.54 |
| Phasic spiking | 6.29 | 2.36 | 1.22 |
| Tonic bursting | 7.24 | 4.75 | 2.46 |
| Phasic bursting | 5.58 | 2.16 | 1.23 |
| Mixed mode | 4.36 | 3.36 | 1.11 |
| Spike frequency adaptation | 4.38 | 4.57 | 1.49 |
| Class 1 | 5.55 | 2.16 | 1.11 |
| Class 2 | 7.19 | 3.87 | 2.92 |
| Spike latency | 6.41 | 4.12 | 1.36 |
| Subthreshold oscillations | 7.31 | 2.22 | 1.25 |
| Resonator | 4.78 | 4.36 | 1.72 |
| Integrator | 5.97 | 3.75 | 1.59 |
| Rebound spike | 7.94 | 2.42 | 2.26 |
| Rebound burst | 6.71 | 4.56 | 2.49 |
| Threshold variability | 5.26 | 2.35 | 1.32 |
| Bistability | 7.44 | 3.21 | 1.79 |
| Mean Value | 6.09 | 3.35 | 1.67 |

## VII. DESIGN AND HARDWARE IMPLEMENTATION

This section presents a hardware implementation structure for the proposed PWL neuron models. To evaluate accuracy and capability of the models, the proposed hardware utilizes full shape signaling of the neurons, however, the proposed models can be implemented with lower cost for spike timing approaches based on Address Event Representation (AER) communication bus [5],[6] in which signal shaping is not concerned. In terms of learning algorithm, both supervised and unsupervised training implementation can be used as depicted in Fig. 9-a.

The proposed architecture consists of three major sub-blocks analogous to different parts of a generalized neural network, including: neuron model, synaptic weights and the training mechanism. For every neuron, each input is multiplied by its pre-synaptic weight and added to the other inputs to provide the total input current of the neuron, where the weights change based on the learning mechanism. In the output of a neuron, the firing time is determined based on the corresponding input and the training algorithm. If it is required in the training mechanism, neuron firing time can be calculated using a hardware counter. Targeting large scale network simulations in this architecture, several neurons are packed for resource sharing and pipelining. Relying on a recursive approach to solve the ordinary differential equations (ODEs) of the neuron model, a recursive structure for each part of the neuron is utilized. In fact, in each clock pulse, one neuron receives input values and calculates input current of the neuron then runs neuron model once and applies training algorithm. Fig. 9-b shows this structure with more details. The mechanism of each section is explained as follows.

### 1. W Unit

In the proposed structure some of the principle synaptic characteristics like weight changing are considered. However this structure is flexible to add more sophisticated synaptic dynamics, like STDP learning. This unit stores the synaptic weights (Ws) and the changing of their values in the training phase. Moreover, it computes the value of the input currents for neurons using Ws. Detailed structure of this unit is shown in Fig. 10. This unit is divided into two subsections:

#### 1.1. Weights Bank

The first one is a storage buffer for Ws called Weights Bank. As Fig. 10-a shows, this subsection includes M buffers

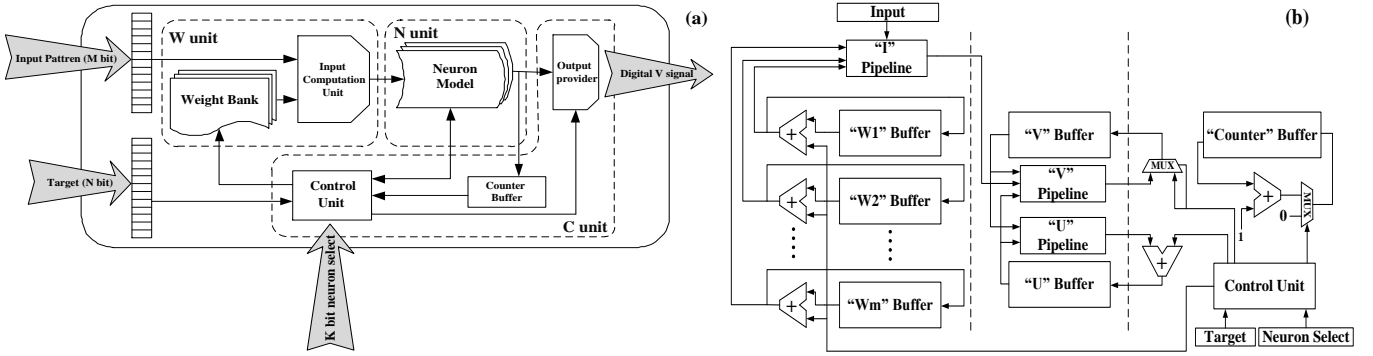

Fig. 9. The proposed architecture of the system a) System Block Diagram b) Neuron Array Block Diagram.

(each buffer includes N weights) for M wb-bit inputs to the network. Each buffer can store N the values of Ws for every neuron. The values of Ws in each buffer is shifted as a memory unit per clock pulse and the output Ws of the buffer is entered to the input of that buffer after applying weight changes based on learning rules. Each weight is a digital wb-bit number, where wb can be determined according to the range of the weights needed for a special application. W_change is a wb-bit number which contains the weight changes. This number is provided by control unit. $i_1$, $i_2$ … $i_M$ are wb-bit weighted inputs which are sent to the input computation unit.

### 1.2. Input computation unit

The second subsection is the input computation unit, which is shown in Fig. 10-b in details. This block is responsible for computing the input current (I) for the neurons considering the value of the input weights which are sent from the Weights Bank unit as $i_1, i_2 … i_M$ and M bit input neurons. In the first stage of this unit input weights are multiplied by input neuron values ($C_1, C_2, …, C_M$). If $C_i=1$ the corresponding input weight passes this stage unchanged and if $C_i=0$, 2's complement of the input weight appears in the output of this stage. In the subsequent stages, the input values are calculated in a pipeline structure. Then results are added to the $i\_bias$ to provide input current of the output neuron ($I\_in$). $i\_bias$ is the minimum required current for neurons to guarantee firing and the input current determines rate of spikes. The last stages are delay stages which are determined according to the number of implemented neurons. Structure of this unit consists of I_S stage pipeline for computing $I\_in$ and D_S stage delay for synchronization.

### 2. N Unit:

As it is shown in Fig. 11, this unit consists of digital implementations of neuron models of the proposed PWL models using four sub-blocks namely V_pipeline, U_pipeline, V_buffer and U_buffer. V_pipeline unit includes the computational structure of the **v'** equation, which is implemented as V_S stage pipeline. U_pipeline unit includes the computational structure of the **u'** equation, which is implemented as U_S stage pipeline. Further, V_buffer and U_buffer are storage buffers for values of V and U with storing capacities of V_buffer_size and U_buffer_size for V and U, respectively. Each buffer shifts a memory unit per clock pulse.

In this structure V and U are $v_b$ and $u_b$-bit fixed-point numbers where $v_b$ and $u_b$ are determined according to the range of V and U values and the required accuracy. VO is the output membrane potential of the neuron which is sent to the control unit to be compared with the threshold condition. "Firing" is a one-bit signal sourced from control unit. Control unit sets this bit when VO reaches to the specified threshold and resets it otherwise. The threshold condition (auxiliary equation in the neuron model) is applied to the output values of the V_pipeline and U_pipeline and results are connected to the inputs of the V_buffer and U_buffer to store the new values. To create the recursive relationship, the outputs of the V_buffer and U_buffer are connected to the V_pipeline and U_pipeline units respectively. In fact, in this structure the computational units are shared between some neurons through pipeline and buffers. For correct operation of the pipeline chain W, V and U values, which appear in the outputs of the W unit and the N unit, must be synchronized to belong to the same neuron in each clock pulse.

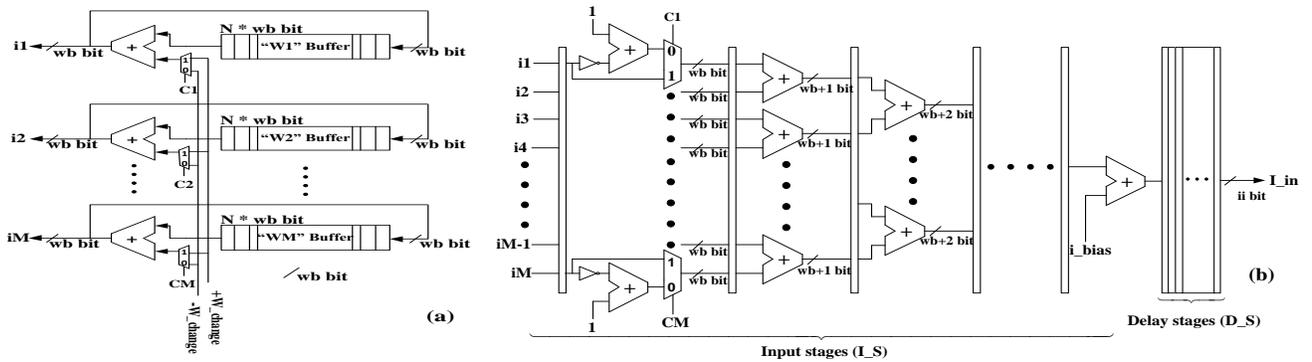

Fig. 10. General structure of the W Unit a) Weights Bank b) Input computation unit.

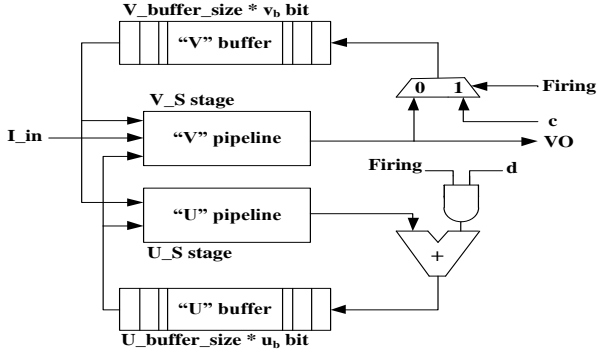

Fig. 11. General overview of N unit.

The required conditions for this synchronization are:

$$\begin{cases} N = V\_buffer\_size + V\_S = U\_buffer\_size + U\_S \\ V\_buffer\_size = U\_buffer\_size \\ V\_S = U\_S \end{cases} \quad (11)$$

where N is the number of implemented neurons. To satisfy the third condition we use delay stages in the output of U_pipeline because this module has less stages than V_pipeline. Regarding hardware implementation, the required condition for updating the weights which their corresponding outputs at the same clock pulse are located in the output of the V_pipeline is:

$$I\_S + D\_S + V\_S = N \quad (12)$$

Considering the number of required neurons, which is fixed and number of computational stages for V and I, we can choose appropriate delay (D_S) to satisfy this equation. In this structure the number of implemented neurons is (D_S = 0):

$$I\_S + V\_S = N \quad (13)$$

For digital implementation, the original continuous time Izhikevich equations are discretized using Euler method. It is well known that replacing multiplication with 'shift' and 'add' operations can result in a considerable cost reduction in digital

implementation. With this motivation, coefficients of the discretized Izhikevich model are chosen as:

$$\begin{cases} v[n+1] = v[n] + dt \cdot (\frac{1}{32} v^2[n] + 4v[n] + 140 - u[n] + I[n]) \\ u[n+1] = u[n] + dt \cdot (a(bv[n] - u[n])) \end{cases} \quad (14)$$

These equations model the same system behavior as (1). In addition, in equation 14, the constants $a$ and $b$ must also be slightly modified from the values of the original Izhikevich model. As an example, a tonic spiking neuron model has been implemented. The modified $a$ and $b$ constants for this neuron type are considered as: $0.203125=1/8+1/16+1/64$ and $0.3125=1/4+1/16$ respectively. The V and U pipelines for the original model of Izhikevich, in the tonic spiking type, and by digitalized $a$ and $b$ and $dt$ constants are shown in detail in Fig. 12 ($a$, $b$, and $c$). The arithmetic operations in equation 14 are assigned to the arithmetic functional units and arranged according to the standard algebraic order of the operations. Data flows directly through the computational tree from the input to the output for all neuron models as shown in Fig. 12. The arithmetic trees maximize the parallelism in time (pipelining) and space (parallel arithmetic units). Computations in each arithmetic tree are fully pipelined to support maximum throughput. Three specific optimizations are made on the algorithm. First, the constant coefficients 4 and 1/32 in equation 14 are implemented as static shift operations (2's complement fixed-pint arithmetic) and use dt=1/ (16*1024) that can be implemented by arithmetic shift and add operations. Second, since multipliers are expensive resources in digital implementation, the multiplication of the parameter 'a' in equation 14 is implemented using a shift and add/subtract operation. This limits the resolution of the values that 'a' can take; however it is efficiently implemented in the hardware. In addition, significant implementation advantages can be gained if $v^2$ could be eliminated in the last model. With this motivation, we modified the original Izhikevich neuron model into PWL models as discussed. This procedure has been repeated for the other PWL models as shown in Table V.

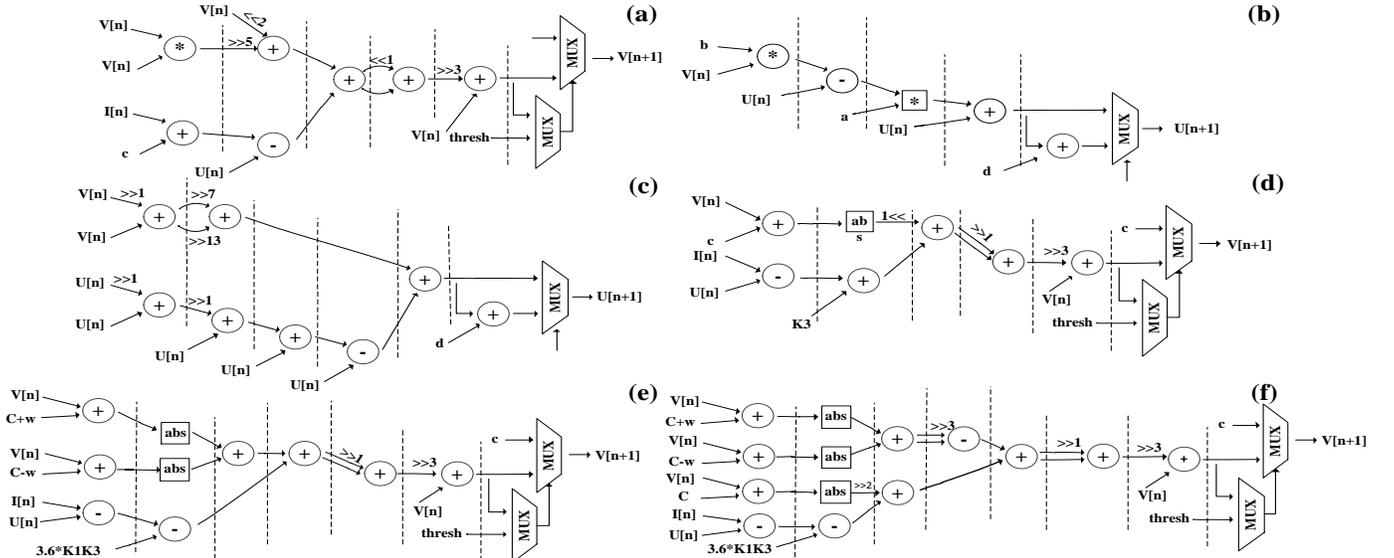

Fig. 12. Arithmetic Pipelines (a) 'v' pipeline in original model (b) 'u' pipeline in original model (c) final modified and digitalized 'u' pipeline for all of the models (d) 'v' pipeline in two piece-wise approximations (e) 'v' pipeline in three piece-wise approximations (f) 'u' pipeline in three piece-wise approximations.

The building blocks of each PWL model and original model are: Multiplier, Adder, Multiplexer and Delay Unit (Flip-Flops or Registers). The parameters that lead us to choose one of these structures are critical path of the circuits, complexity (number of adders and multipliers which required) and the required computational accuracy (word-length). The number of minimum required resources, which are indicated in the scheduling sequence of V unit for implementation of each model, are shown in Table IV.

TABLE IV: The number of minimum resources in the scheduling of V

| Resources | Orig. Model | 2PWL | 3PWL | 4PWL |
|---|---|---|---|---|
| Adder | 6 | 6 | 8 | 11 |
| Multiplier | 1 | - | - | - |
| Multiplexer | 2 | 3 | 4 | 5 |

The critical stage which determines the operation frequency of the system clock in each implementation model is:

$$\begin{cases} T_{\text{Orig.model}} = T_{\text{MUL}} \\ T_{\text{Pwlmodels}} = T_{\text{ADD}} \end{cases} \qquad (15)$$

So the PWL models are expected to work in a higher frequency compared to the original model.

### 3. C Unit

This unit produces the required controlling signals for training the network and includes three sub-blocks: counter buffer, control unit and output provider. Detailed structure of this unit is depicted in Fig. 13.

#### 3.1 Counter buffer

This block is a storage buffer which stores spikes' timing in each neuron in terms of number of clock pulses. This buffer has the capacity of storing N counter values corresponding to N implemented neurons. The stored values in the buffer are shifted in each clock pulse and the output of the buffer is returned to its input. In each storing step the threshold condition (auxiliary equation in Izhikevich model of 2) is checked by control unit and counter value resets to zero if neuron fires and increases one unit otherwise. It should be noted that before resetting counter its value is used by control unit to evaluate suitable weight change during the training phase.

#### 3.2 Control Unit

This is the block that obtains necessary data and applies control signals to other units. Input data to this block are:

− Output V from N unit (VO) for threshold condition checking.
− Output counter value from counter buffer for evaluating weight changes.

− N bit target value from user for evaluating target firing time (or desired time of spike firing).
− "Valid" that is input signal from user which shows input neurons value, target and neuron_select.
− Input neurons which are used in training mechanism if an unsupervised training algorithm has been applied.
− K-bit neuron_select from user for putting V value of desired neuron into the output register.

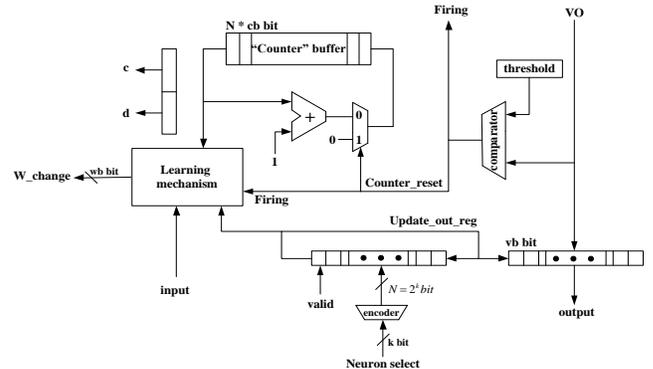

Fig. 13. Control unit of the proposed structure.

This unit contains an internal encoder and logical shift register for producing proper command to the output provider to represent the selected neuron output. Encoder receives k-bit neuron_select number and provides a $2^k$-bit number which contains one bit "1" and other bits "0". When the neuron_select number is valid, user sets "valid" input bit and encoded neuron_select is stored in a $2^k$-bit shift register. When the user resets "valid" bit, shift register that contains $2^k$-1 bit "0" and one bit "1" shifts logically each clock pulse. Therefore update_out_reg signal is one clock pulse in "1" state and N-1 clock pulse in "0" state. This signal is used as write enable signal in the output provider register. The output control signals of this block are:

− V reset command and U change command to N unit.
− Command to output provider for representing desired output
− weight_change: weight change value to Weight Bank block. This value is calculated in the training mechanism module according to the applied training algorithm.

#### 3.3 Output provider

This sub-block is a register which provides V at the output. When the V value of the selected neuron appears in the output of the N unit, the control unit sends a command to the output provider to replace its register current value with the new one. When the V of the other neurons appears in output of N unit, the output provider register keeps its value. In fact, this register updates once in every N clock pulse.

TABLE V: Discrete equations of "V" for all PWL models, the digitalized K parameters and number of stages in pipeline implementation.

| Models | Discrete "V" equation | Parameters | Pipeline Stages |
|---|---|---|---|
| 2 PWL | $v[n+1] = v[n] + dt(k1\|v[n] - 62.5\| + k2 - u[n] + I[n])$ | K1=0.75=1/2+1/4, K2=20 | 5 |
| 3 PWL | $v[n+1] = v[n] + dt(k1(\|v[n] + 62.5 + k2\| + \|v[n] + 62.5 - k2\|) - k3k2k1 - u[n] + I[n])$ | K1=0.625=1/2+1/8, K2=5.8, K3=6.4 | 6 |
| 4 PWL | $v[n+1] = v[n] + dt(k2(\|v[n] + 62.5 + k3\| + \|v[n] + 62.5 - k3\|) + k1\|v[n] + 62.5\| - 4k2k3 - u[n] + I[n])$ | K1=0.375=1/4+1/8, K2=0.75, K3=11 | 7 |

## VIII. Character Recognition and Training Algorithm

Neural network approach is a strong tool for modeling data, based on computer training, which are basically trained to perform complex functions in various fields of applications including pattern recognition, identification, classification, speech, vision and control systems [40]. As a case study we utilized a two layers spiking neural network for character recognition based on the Izhikevich and PWL models. In this implementation the tonic spiking neurons are used for student rule because of two important reasons: The first one is minimum distortion in its initial behavior signal (V) and the second one is its minimum bursting behavior. These two ideal characteristics make tonic spiking as a suitable neuron model for our proposed training algorithm in the case studies. Numerical values are constrained to a 20-bit fixed-point (8.12) representation. The structure of the proposed network is depicted in Fig. 14. In this model, there are two layers, the first layer acts as input neurons using a rate-based coding to recognize the patterns. Input patterns were presented to the first layer of neurons (which we refer to as level 1), with each pattern pixel corresponding to a separate input neuron. Thus the number of neurons in level 1 was equal to the number of pixels in the input pattern. The number of second layer neurons (which we refer to as level 2) was equal to the number of training patterns (i.e. 26). This is because each level 2 neuron tunes itself to the valid frequency firing rate only if it recognizes its assigned pattern. The input layer neurons are fully connected to the layer 2 neurons with synapses for which weights are determined by rate-based coding as shown in Fig. 14. The output layer receives spikes from the input during the training stage. Also the excitation value (I) for every output Izhikevich neuron is considered as:

$$I_j = \sum_{i=1}^{i=M} I_i \times W_{ij} + I_0 \tag{16}$$

where M is the number of input neurons, I is bipolar coded, W is the weighting factors and $I_0$ is the bias for putting the neuron in firing state.

Neurons, which are processing elements in the network, are connected to each other through a set of weights. These weights are adjusted based on an error-minimization technique called back-propagation rule. The weights change when the corresponding output neuron fires. This algorithm is derived to minimize the error in the output spike rate through gradient decrement in the W space. This error can be measured as:

$$E = ((Counter_j - t_j))^2 \tag{17}$$

where j is the number of neurons in the output layer, t is the period time of the target frequency that the $j^{th}$ output neuron trace and counter is the period time of the output frequency of $j^{th}$ neuron. The gradient of the error function is given by:

$$\frac{\partial E}{\partial W_{ij}} = 2(Counter_j - t_j) . \frac{\partial Counter_j}{\partial W_{ij}} \tag{18}$$

In tonic spiking neuron the relation between the input current (I) and the counter is negative linear as [35]:

$$Counter_j \propto -I_j \tag{19}$$

With substitution equation (16) and (19) we have:

$$\frac{\partial Counter_j}{\partial W_{ij}} = -k \frac{\partial I_j}{\partial W_{ij}} = -k^. I_i \tag{20}$$

where k and $k^.$ are positive constants. Finally with substitution equation (17) and (20) weight change policy can be written as:

$$W_{ij}(k+1) = W_{ij}(k) + \Delta W_{ij} \tag{21}$$

with:

$$\Delta W_{ij} = -\mu \frac{\partial E}{\partial W_{ij}} = \alpha \times (I_i \times (Counter_j - t_j))$$

where $\alpha$ is a positive coefficient obtained by multiplying $k^.$ and $\mu$. $\Delta W$ is the weight change, i is the number of neurons in the input layer, $\alpha$ is a constant coefficient for normalization of the weight changes. I is the input pixel which is considered as bipolar coding, it means that if the input pixel is white we have -1 and +1for black pixel. As it is mentioned before we defined two high (80 Hz) and low (10 Hz) frequencies for this purpose. When the output neuron recognizes the character correctly, the output frequency of the neuron traces high and the other neurons trace low frequency. This network can be used for any high resolution pattern recognition.

The proposed supervised training algorithm is implemented as shown in Fig. 15, which should replace the training mechanism module in Fig. 13. In this structure when the target number is valid, the user sets 'Valid' input bit and the target number will be stored in the target register. When the user resets 'Valid' bit, target register shifts logically each clock pulse. $t_N$ is a one bit signal which is connected to the last bit of the target register and contains logical output of the selected neurons in each clock pulse. This signal is used as "Select" signal for a multiplexer, providing the required spike rate for the corresponding neuron. The difference of the desired firing time and the real firing time is evaluated by a subtractor and result is shifted mathematically to implement alpha coefficient in equation 20. This applies if neuron fires a spike ("Firing" signal is equal to "1"). The time step is 1/(16*1024) second.

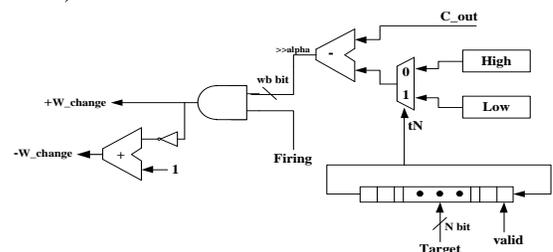

Fig. 15. Proposed supervised training algorithm

As it is discussed above, the teacher neuron is supposed to be a tonic spiking neuron model. Let $\tilde{V}$ and $\tilde{Y}$ denote the teacher's membrane potential and its binary spike train, respectively. The teacher accepts different frequency ranges in the output, which two high and low frequencies are considered as 'valid' and 'invalid' inputs respectively.

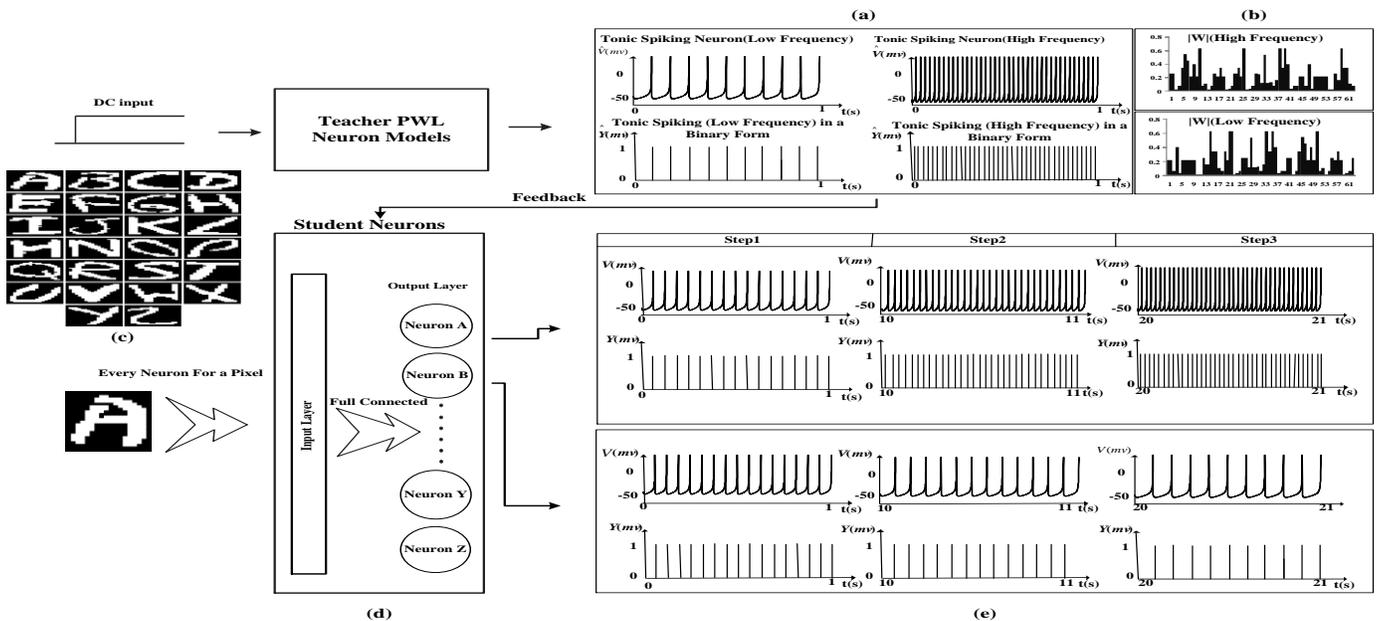

Fig. 14. A case study with rate-based coding train for 1D coordinate transform on FPGA.(a) The teacher neuron. If the neuron is trained well it should mimic the behavior of the teacher neuron. (b) The weight changing after training phase for recognized pattern (high frequency rate) and unrecognized pattern (low frequency rate). (c) Some input patterns from MNIST database (d) Proposed structure for the network (e) Training steps for output neurons.

In this network all output neurons fire with the 'invalid' frequency except one which corresponds to the input character. Here four student classes are assumed, which the input-output relationship in these students are Izhikevich, 4PWL, 3PWL and 2PWL models. Students trace the spiking frequency of the output teacher by adjusting their weight parameters where models ($a$, $b$, $c$, $d$, $k_1$, $k_2$, $k_3$) are fixed. The training process in three time steps is depicted in Fig. 14. This Figure shows the convergence speed in different student models. In this figure the character 'A' is applied to the network which 'neuron A' traces high and the other neurons (in the figure only B is depicted as an example) trace low frequency. Fig. 14-b shows synaptic weights distribution after training.

## IX. IMPLEMENTATION RESULTS

To verify the validity of the models and the architecture, the case study explained in VIII is implemented on a XILINX XUP Virtex-II Pro Development System, which provides a hardware platform that consists of a high performance Virtex-II Pro XC2VP30 Platform FPGA surrounded by a comprehensive collection of peripheral components. Fig. 16 shows oscilloscope photographs of the dynamical behavior of a single neuron implemented on this FPGA platform using Izhikevich and the three proposed PWL models for the three different test cases: tonic spiking, tonic bursting and trained signal. All three test cases are in response to a step signal (at 1/ (16*1024) sec). Utilized parameters for the four test cases were obtained from Izhikevich [4] and the modified models explained before. The device utilization for implementation of 30 neurons based on the Izhikevich and the proposed PWL models are summarized in Table VI.

For a fair comparison, the original Izhikevich model was implemented by three different types of fixed-point multipliers to find the most efficient implementation. The results show that the implemented PWL models are significantly faster than Izhikevich's (approximately 9.2 times on a Virtex II Pro) with simple combinational multiplier. This was expected because the original model has a $v^2$ term which leads to a longer critical path in the circuit. In addition, since Izhikevich model requires high performance multipliers, the number of neurons is limited to the number of embedded multipliers in the FPGA. If we use a full pipelined multiplier, the clock pulse frequency of Izhikevich model becomes almost equal to PWL models but requires a considerably higher area and resources compared with the PWL models. If we use a booth serial multiplier, we can solve the problem of high usage of area and resources and keep the clock frequency of FPGA almost equal to PWL models, but as seen in Table VI, overall clock frequency of the system is divided by 11 because n-bit booth multiplier needs n/2+1 clock pulses to calculate the result. Here we use 20-bit booth multiplier that needs 11 clock pulses to provide the result and it results in the overall clock pulse of the system 11 times slower than FPGA clock pulse.

For evaluating performance of the proposed models in a network for a pattern recognition task, the widely studied case of standard handwritten alpha-digits recognition of [41] is implemented. The MNIST database is used, which contains handwritten binary 20×16 digits of 'A' to 'Z' by 39 writers. This database has been used as a test bench for many learning algorithms. In order to achieve learning, 29 different handwrites from each character 25 times are fed to the network for training. After this phase, for testing the network ten remained handwrites (for each character) are presented to the network. Results show 91.7% accuracy in the recognitions, where a similar case study has reported 95% accuracy for a traditional artificial neural network with back-propagation learning algorithm, with 300 hidden neurons [42]. Interestingly, our network is based on SNN and the result has been obtained from hardware implementation.

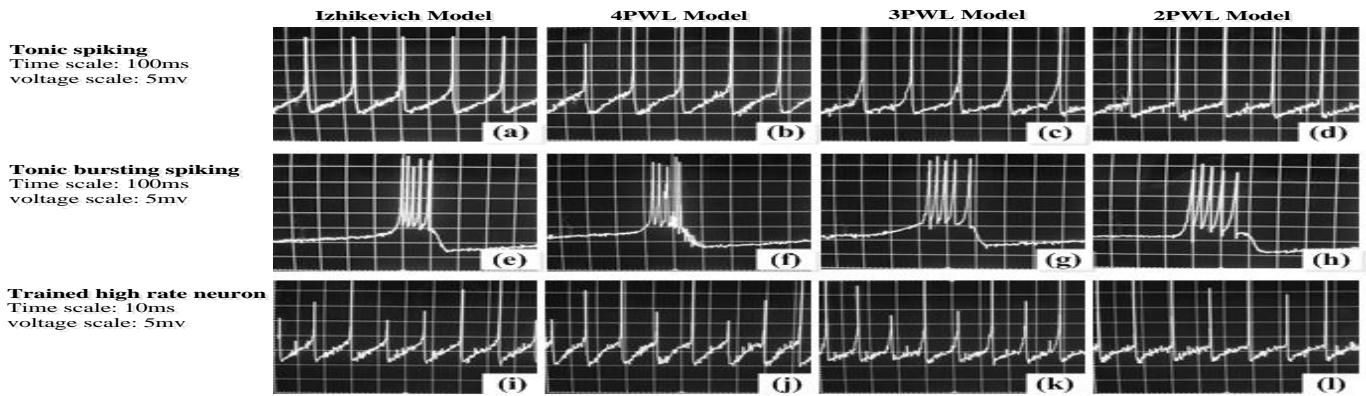

Fig. 16. Comparison between output of the Izhikevich model and the PWL models implemented on XILINX Virtex-II Pro XC2VP30. Signals are physically produced and observed on the oscilloscope. (a) Tonic spiking implementation, Izhikevich model. (b) Tonic spiking implementation, 4PWL model. (c) Tonic spiking implementation, 3PWL model. (d) Tonic spiking implementation, 2PWL model. (e) Tonic bursting spiking implementation, Izhikevich model. (f) Tonic bursting spiking implementation, 4PWL model. (g) Tonic bursting spiking implementation, 3PWL model. (h) Tonic bursting spiking implementation, 2PWL model. (i) Trained high rate neuron, Izhikevich model. (j) Trained high rate neuron, 4PWL model. (k) Trained high rate neuron, 3PWL model. (l) Trained high rate neuron, 2PWL model.

TABLE VI. Device utilization of the XILINX Virtex-II Pro (A) Based on speed optimization goal (B) Based on area optimization goal.

|   | Resource | Izhikevich Model (Combinational MUL.) | Izhikevich Model (Full Pipelined MUL.) | Izhikevich Model (Booth MUL.) | The 2PWL | The 3PWL | The 4PWL | Total Available |
|---|---|---|---|---|---|---|---|---|
| **A** | Slice FF's | 432 | 1294 | 510 | 374 | 450 | 493 | 27392 |
|   | RAM (Byte) | 164 | 245 | 164 | 150 | 155 | 158 | 54784 |
|   | 4-LUTs | 1176 | 1009 | 665 | 453 | 520 | 617 | 27392 |
|   | Max Speed | 26.503Mhz | 241.937Mhz | 15.72Mhz | 241.937Mhz | 241.937Mhz | 241.937Mhz | 400Mhz |

|   | Resource | Izhikevich Model (Combinational MUL.) | Izhikevich Model (Full Pipelined MUL.) | Izhikevich Model (Booth MUL.) | The 2PWL | The 3PWL | The 4PWL | Total Available |
|---|---|---|---|---|---|---|---|---|
| **B** | Slice FF's | 424 | 1266 | 498 | 365 | 441 | 491 | 27392 |
|   | RAM (Byte) | 142 | 245 | 164 | 150 | 155 | 158 | 54784 |
|   | 4-LUTs | 1173 | 1009 | 626 | 441 | 500 | 602 | 27392 |
|   | Max Speed | 26.503Mhz | 234.102Mhz | 13.03Mhz | 204.311Mhz | 204.311Mhz | 204.311Mhz | 400Mhz |

## X. CONCLUSIONS

This paper presents a set of multiplier-less biologically inspired neuron models based on well-known model of Izhikevich. Findings show that these modifications simplify the hardware implementation yet demonstrate similar dynamical behavior. Since the models consist of simple arithmetic operations (addition/subtraction and shift) a large number of neurons can be implemented on FPGA without any need to the embedded multipliers on the FPGA chip. Architecture has been developed for a network of the proposed neuron models which can be implemented in a high speed and low area. The PWL models have reached approximately 200 MHZ clock rate (almost 9.2 times faster compared with the original Izhikevich model on a Virtex II Pro) with a comparable area or a 2-4×area improvement over the fully pipelined multiplier model at a comparable operating frequency. It is clear that if the operating frequency increases, more virtual neurons can be implemented on a fixed number of physical neurons with the same sampling time (dt=1/(16*1024)).